\documentclass[10pt,twocolumn,letterpaper]{article}

\usepackage[pagenumbers]{wacv}      

\usepackage{siunitx}
\usepackage{multirow}
\usepackage{dblfloatfix}
\usepackage{makecell}

\newcommand{\rev}[1]{\textcolor{black}{#1}}

\definecolor{wacvblue}{rgb}{0.21,0.49,0.74}
\usepackage[pagebackref,breaklinks,colorlinks,allcolors=wacvblue]{hyperref}

\def\wacvPaperID{2132} 
\def\confName{WACV}
\def\confYear{2027}

\title{ThermalSplat: Generalizable Pose-Free Gaussian Splatting for Sparse-View Thermal 3D Reconstruction}

\author{
Prasanna S \quad
Gopi Raju Matta \quad
Divya Madhuri Vemunuri \quad
Kaushik Mitra \\[2pt]
Indian Institute of Technology Madras \\[4pt]
{\tt\small s.prasann03@gmail.com,\;
\{ee17d021,ee24d004\}@smail.iitm.ac.in,\;
kmitra@ee.iitm.ac.in}
}

\begin{document}
\maketitle
\begin{abstract}
Thermal infrared imaging is vital for applications like search-and-rescue and autonomous navigation, yet translating it into 3D scene representations remains a critical bottleneck. The low-texture nature and view-dependent emissivity of thermal images cause standard pose estimators to fail catastrophically, collapsing standard 3D Gaussian Splatting pipelines under sparse-view constraints. Furthermore, existing thermal reconstruction methods rely heavily on dense capture and computationally expensive per-scene optimization. Critically, many of these approaches depend on aligned RGB-thermal datasets to obtain reliable camera poses and initial point clouds, rendering them unusable in purely thermal, unposed environments. To break this bottleneck, we present ThermalSplat, the first generalizable, feed-forward Gaussian splatting framework that reconstructs 3D scenes directly from sparse thermal images without requiring known camera poses, RGB guidance, or per-scene training. We overcome thermal-specific challenges using a \emph{soft census loss} for brightness-invariant structural matching and a \emph{boundary structure loss} to explicitly supervise temperature discontinuities. Trained on an aggregated multi-source dataset, ThermalSplat synthesizes sharp reconstructions of unseen scenes in a single feed-forward pass. Extensive experiments demonstrate that ThermalSplat holds a strict advantage over scene-specific thermal splatting methods in the sparse regime. Furthermore, even compared against these baselines operating under their ideal dense capture settings, ThermalSplat achieves superior perceptual fidelity. By eliminating the need for computationally expensive per scene optimization, ThermalSplat offers an efficient and generalizable approach to thermal 3D reconstruction.
\end{abstract}    
\section{Introduction}
\label{sec:intro}

\begin{table}[t]
\centering
\caption{Motivation for pose-free thermal reconstruction. We evaluate state-of-the-art Structure-from-Motion (SfM) methods on our thermal test scenes by measuring the number of successfully registered images for pose estimation and the resulting extracted 3D points. While dense inputs yield partial success, these methods fail to extract reliable 3D points in the sparse (2-view) setting, highlighting the fundamental necessity of our pose-free framework.}
\label{tab:sfm}
\resizebox{\columnwidth}{!}{%
\begin{tabular}{@{}ll rrrrrr@{}}
\toprule
& & \multicolumn{6}{c}{\textbf{Thermal Scenes}} \\
\cmidrule(l){3-8}
& & Hand & Dark & Fire3 & Tower & UAV6 & Bldg. \\
\midrule
\multicolumn{2}{c}{\textbf{Total Images}} & \textbf{42} & \textbf{86} & \textbf{41} & \textbf{177} & \textbf{132} & \textbf{96} \\
\midrule
\textbf{Method} & \textbf{Metric} & \multicolumn{6}{c}{\textit{Dense Context (All Views)}} \\
\midrule
\multirow{2}{*}{\textbf{COLMAP}}
& Registered Images & 42 & 0 & 0 & 11 & 132 & 35 \\
& 3D Points   & 622 & 0 & 0 & 493 & 22,209 & 222 \\
\addlinespace
\multirow{2}{*}{\textbf{VGGT}}
& Registered Images & 42 & 0 & 0 & 0 & 132 & 96 \\
& 3D Points   & 100,000 & 0 & 0 & 0 & 381 & 3,515 \\
\addlinespace
\multirow{2}{*}{\textbf{VGGSfM}}
& Registered Images & 42 & 80 & 4 & 133 & 110 & 96 \\
& 3D Points   & 3,011 & 12,459 & 1,343 & 11,431 & 4,677 & 5,278 \\
\midrule
& & \multicolumn{6}{c}{\textit{Sparse Context (2 Views)}} \\
\midrule
\multirow{2}{*}{\textbf{VGGT}}
& Registered Images & 0 & 2 & 0 & 0 & 0 & 0 \\
& 3D Points   & 0 & 76,691 & 0 & 0 & 0 & 0 \\
\addlinespace
\multirow{2}{*}{\textbf{VGGSfM}}
& Registered Images & 0 & 0 & 0 & 0 & 0 & 0 \\
& 3D Points   & 0 & 0 & 0 & 0 & 0 & 0 \\
\bottomrule
\end{tabular}%
}
\end{table}

Thermal infrared imaging has become an indispensable sensing modality across a wide range of real-world applications, from search-and-rescue operations and building energy diagnostics to infrastructure inspection and autonomous navigation under adverse lighting and weather conditions. Unlike visible-light cameras, thermal sensors capture emitted radiance rather than reflected light, enabling perception in complete darkness, smoke, and fog. Yet despite this practical utility, translating thermal imaging into robust 3D scene representations remains a largely underexplored and highly challenging bottleneck.

The challenge is fundamental. Thermal images are inherently low-texture and low-contrast. Scenes appear as smooth gradients of emitted heat, with object boundaries defined by temperature discontinuities rather than the rich surface texture and colour variation that visible-light reconstruction pipelines rely upon. Spatial resolution is typically lower than RGB cameras, and standard feature detectors and matchers trained on visible imagery transfer poorly to the thermal domain~\cite{usamentiaga2014infrared}. As demonstrated in Table~\ref{tab:sfm}, due to these properties and the severe view-dependent emissivity, classical Structure-from-Motion (SfM) pipelines like COLMAP~\cite{schonberger2016structure}, and even recent feed-forward pose estimators like VGGT~\cite{wang2025vggt}, fail catastrophically at extracting reliable camera poses or initial point clouds under sparse capture conditions. Consequently, standard multi-view reconstruction and 3D Gaussian Splatting pipelines collapse when applied to sparse thermal inputs.

While sparse-view 3D reconstruction has seen remarkable progress in the visible-light domain through feed-forward Gaussian splatting methods~\cite{chen2024mvsplat,charatan2024pixelsplat,huang2026none}, these models are trained exclusively on large-scale RGB datasets. Applying them zero-shot to thermal inputs produces perceptually degraded reconstructions that blur critical temperature boundaries and produce inconsistent intensity fields across novel views. Concurrently, the thermal reconstruction literature has successfully modelled emitted radiance through methods such as PhysIR~\cite{gao2026physir}, ThermalGaussian~\cite{lu2024thermalgaussian}, Thermal3D-GS~\cite{chen2026thermal3d}, ThermoNeRF~\cite{hassan2025thermonerf}, and Thermal-NeRF~\cite{lin2024thermalnerf}. However, a critical limitation unifying all of these thermal methods is their reliance on dense multi-view capture and computationally expensive per-scene optimization. Crucially, to bypass the low-texture nature of thermal images, most of these methods typically depend on aligned RGB-thermal datasets to obtain reliable Structure-from-Motion (SfM) initializations and camera poses. For instance, models like ThermalGaussian actively rely on paired RGB images, often utilizing composite formats like MSX, to extract poses and provide rich texture guidance. Because they depend on visible-light data to bootstrap the underlying geometry, these existing works fail in purely thermal, unposed operational environments. None addresses the sparse-view setting in a generalizable, pose-free manner.

To break this bottleneck, we ask: \textit{can a generalizable, feed-forward sparse-view reconstruction framework be made to work reliably in the thermal domain without known poses, RGB guidance, or per-scene training?} We answer affirmatively with \textbf{ThermalSplat}, the first generalizable, feed-forward Gaussian splatting framework for sparse-view thermal 3D reconstruction. Built on NAS3R~\cite{huang2026none}, a large pose-free multi-view geometry backbone, ThermalSplat is adapted to the thermal modality through three complementary contributions. First, we construct an aggregated multi-source thermal training corpus from five publicly available datasets spanning indoor, outdoor, and aerial scenes. Second, to address the blurring of structural boundaries, we introduce a \emph{boundary loss} that explicitly preserves the temperature discontinuities that constitute the primary geometric cues in thermal imagery. Third, to resolve the intensity shifts caused by view-dependent emissivity and sensor auto-ranging, we introduce a \emph{census loss} that preserves local intensity relationships despite radiometric variations.

Extensive evaluations on held-out thermal scenes demonstrate that ThermalSplat synthesizes sharp, multi-view consistent reconstructions of entirely unseen scenes in a single feed-forward pass. By combining thermal-specific adaptation with tailored structural supervision, our approach significantly surpasses existing generalizable frameworks fine-tuned on thermal data. More importantly, ThermalSplat establishes a strict advantage over scene-specific thermal Gaussian splatting methods in the sparse regime, and achieves superior perceptual fidelity even when these baselines are provided with their ideal dense capture settings, delivering exceptionally faithful thermal representations.

\begin{itemize}
  \item We introduce \textbf{ThermalSplat}, the first generalizable feed-forward framework for sparse-view thermal 3D reconstruction. It successfully eliminates the need for camera poses, RGB assisted SfM initialization and per-scene optimization, setting a new state-of-the-art standard for the thermal domain.
 
\item We strategically adapt a \textbf{boundary loss} and a \textbf{census loss} to address the geometric and structural challenges unique to thermal imagery. By integrating these structural losses, our framework seamlessly handles view-dependent emissivity and sensor auto-ranging while preserving critical temperature discontinuities without requiring external depth supervision.
  
\item We conduct \textbf{comprehensive experiments} demonstrating that ThermalSplat \rev{generalizes robustly to unseen scenes}. Our method significantly surpasses existing generalizable frameworks fine-tuned on thermal data, and it \rev{clearly} outperforms scene-specific thermal baselines in the sparse regime.
\end{itemize}
\section{Related Work}
\label{sec:related}

\paragraph{Generalizable Gaussian Splatting.}
Classical novel view synthesis pipelines combine Structure from Motion (SfM)~\cite{schonberger2016structure} and Multi-View Stereo to recover camera poses and geometry before optimizing scene-specific representations. Recent advances in Neural Radiance Fields (NeRF)~\cite{mildenhall2021nerf} and 3D Gaussian Splatting (3DGS)~\cite{kerbl20233d} substantially improve rendering quality, but both rely on dense multi-view observations and per-scene optimization, making them brittle under sparse capture constraints. Generalizable feed-forward approaches address this by learning scene priors from large-scale datasets. PixelSplat~\cite{charatan2024pixelsplat} and MVSplat~\cite{chen2024mvsplat} introduced feed-forward Gaussian splatting from image pairs using depth estimation techniques. Recent methods have progressively eliminated the dependence on calibrated camera poses. SPFSplat~\cite{huang2025no} and NAS3R~\cite{huang2026none} achieve fully unposed reconstruction through joint pose prediction and global cross-view reasoning. Despite their success, all existing generalizable methods are trained exclusively on visible-light RGB datasets. Their learned priors and assumed appearance statistics fundamentally fail to transfer to the thermal domain.

\paragraph{Thermal 3D Reconstruction.}
Thermal scene reconstruction has primarily focused on adapting NeRF and 3DGS through scene-specific optimization. ThermalGaussian~\cite{lu2024thermalgaussian} introduced thermal 3DGS by jointly optimizing RGB and thermal representations, while methods like Thermal3D-GS~\cite{chen2026thermal3d} and PhysIR-Splat~\cite{gao2026physir} explicitly model infrared-specific imaging physics to achieve physically consistent rendering. Similarly, ThermoNeRF~\cite{hassan2025thermonerf}, Thermal-NeRF~\cite{lin2024thermalnerf}, and ThermalMix~\cite{ozer2024exploring} explore joint multimodal scene representations, and Wild-Thermal~\cite{aydin2026thermal} addresses radiometric fluctuations through photometric stabilization. However, a critical limitation unifies the current literature: these methods universally rely on dense multi-view capture and computationally expensive per-scene optimization. In contrast, ThermalSplat is the first generalizable, feed-forward Gaussian splatting framework designed explicitly for unposed, sparse-view thermal reconstruction without requiring per-scene training.
\section{Method}
\label{sec:method}

\subsection{Preprocessing}

Thermal images are single-channel grayscale images with scene-dependent radiometric ranges. To adapt them to the RGB-pretrained backbone, each thermal image is replicated across three channels without modifying the network architecture. To reduce \rev{intensity} variations across different thermal cameras and capture conditions, we apply a global \rev{intensity} normalisation using robust percentile bounds computed over the entire training corpus. Each pixel intensity $x$ is normalised as
\begin{equation} \hat{x} = \frac{x - p_\alpha}{p_{1-\alpha} - p_\alpha}, \end{equation}
where $p_\alpha$ and $p_{1-\alpha}$ denote the lower and upper global percentile values\rev{,} with $\alpha=0.01$ \rev{(1st/99th percentiles)}~\cite{Gonzalez2008}. The resulting values are clipped to $[0,1]$

\subsection{Architecture}
ThermalSplat is built upon NAS3R~\cite{huang2026none}, a feed-forward, pose-free
Gaussian splatting framework driven by a large transformer-based multi-view
geometry backbone; we refer to this backbone hereafter simply as the
backbone. Unlike cost-volume methods that require posed context views, the
backbone ingests a set of unposed thermal images and reasons about cross-view
geometry purely through attention, making it well suited to thermal capture
where reliable camera calibration is frequently unavailable.
 
\begin{figure*}[t]
    \centering
    \includegraphics[width=\textwidth]{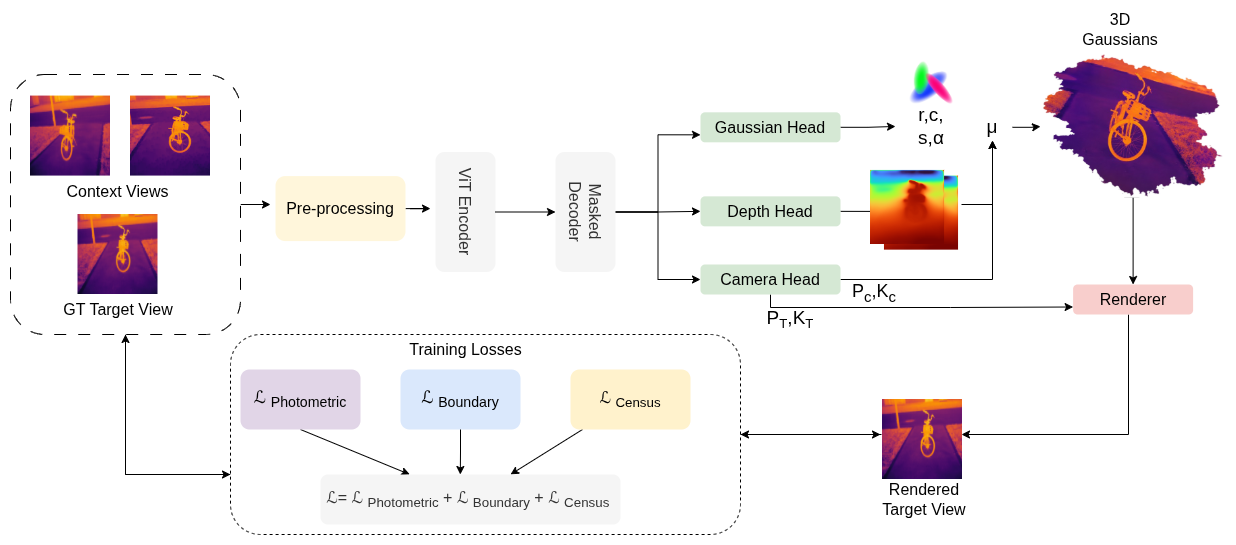}
    \caption{Overview of the ThermalSplat framework. Given a sparse set of unposed thermal context views, the images undergo thermal-specific preprocessing before being tokenized by a Vision Transformer (ViT) encoder and processed by a masked decoder. The multi-view tokens are then routed to three specialized prediction heads: a Camera Head that regresses relative poses, a Depth Head that predicts dense depth maps, and a Gaussian Head that predicts 3D Gaussian parameters. These are unprojected to form a 3D scene representation and rendered into novel target views via differentiable rasterisation in a single feed-forward pass. To overcome the low-contrast and view-dependent emissivity of thermal imagery, end-to-end training is supervised by a photometric loss augmented with two thermal-adapted structural losses: a boundary structure loss ($\mathcal{L}_{\text{boundary}}$) to preserve critical temperature discontinuities and a soft census loss ($\mathcal{L}_{\text{census}}$) for brightness-invariant local intensity matching.}
    \label{fig:framework}
\end{figure*}

\paragraph{Multi-view transformer encoder.}
Each context view is tokenized into patch embeddings by a pretrained vision
transformer encoder~\cite{dosovitskiy2020image}. The resulting per-view token sequences
are processed by an aggregator that alternates between \emph{frame-wise}
self-attention, which models appearance and structure within a single view,
and \emph{global} self-attention across all views, which propagates
correspondence and geometric context between them~\cite{wang2025vggt}. This
alternating-attention design replaces explicit feature matching or
cost-volume construction with learned cross-view reasoning, which is
advantageous in the thermal domain where low texture and reduced contrast
render appearance-based matching unreliable.
 
\paragraph{Prediction heads.}
The aggregated multi-view tokens are decoded by three lightweight heads. A \emph{camera head} regresses the relative pose of each view, and optionally its intrinsics, directly from the global tokens, removing the need for known poses at inference. A \emph{depth head} predicts a dense per-view depth map, from which per-pixel 3D points are obtained by unprojection. A \emph{Gaussian head} predicts the parameters of one 3D Gaussian per pixel (opacity, covariance, and spherical harmonic appearance coefficients) anchored at the unprojected points. The complete set of Gaussians from all context views is rendered into novel target views via differentiable tile-based rasterisation~\cite{kerbl20233d}, and the model is trained end-to-end through the rendering objective of Eq.~\ref{eq:total_loss}.

\subsection{Training Losses}
\label{sec:losses}
Two failure modes motivate our structural loss terms. First, standard photometric losses blur the temperature discontinuities at object boundaries, which are often the most reliable structural cues in thermal imagery. Moreover, thermal intensity is influenced by material properties and imaging conditions, causing gradient-based edge losses to respond to intensity variations rather than underlying scene structure. Second, frame-to-frame intensity changes introduced by the thermal imaging process make pixel-wise supervision unreliable even when the geometry is correctly recovered. We therefore employ a boundary structure loss that emphasizes structural edges and a soft census loss that compares local intensity relationships instead of absolute pixel values. Both losses are self-supervised and require no external annotations.

The total training objective combines a standard photometric loss with the two complementary thermal-oriented structural losses:

\begin{equation}
\mathcal{L}
=
\mathcal{L}_{\text{photo}}
+
\lambda_b \mathcal{L}_{\text{boundary}}
+
\lambda_c \mathcal{L}_{\text{census}},
\label{eq:total_loss}
\end{equation}

where $\mathcal{L}_{\text{photo}}$ denotes the combined pixel ($\ell_2$) and
perceptual (LPIPS) rendering loss, $\mathcal{L}_{\text{boundary}}$ and
$\mathcal{L}_{\text{census}}$ denote the boundary structure and census losses,
while $\lambda_b$ and $\lambda_c$ are scalar weighting coefficients
controlling the contributions of the boundary structure and census losses,
respectively.

\paragraph{Boundary structure loss.}

In thermal imagery, the observed intensity depends not only on scene temperature but also on material properties such as emissivity. As a result, conventional edge losses based on raw image gradients may respond to intensity variations rather than true structural boundaries. We therefore adapt the colour-invariant gradient response of~\cite{zhou2025lita}, normalising image gradients by the local intensity to obtain a boundary response that is substantially less sensitive to local intensity scaling.

Spatial gradients are obtained by convolving each image with Gaussian derivative filters at a fixed scale, giving per-pixel responses $\{f_x,f_y\}$ for the image intensity $f$. Normalising by the local intensity gives the boundary response

\begin{equation}
\mathcal{S}(I)
=
\log\!\left(
\frac{f_x^2+f_y^2}{f^2+\epsilon}
+\epsilon
\right),
\label{eq:boundary_resp}
\end{equation}

which depends on boundary location and relative edge strength but not on absolute intensity: under any local rescaling $f \rightarrow cf$, the factor $c$ cancels in the ratio. The response is subsequently instance-normalised per image to standardise its dynamic range. The loss is the mean absolute difference between the boundary responses of rendered and ground-truth target views:

\begin{equation}
\mathcal{L}_{\text{boundary}}
=
\frac{1}{|\mathcal{T}|}
\sum_{i\in\mathcal{T}}
\left|
\mathcal{S}(\hat{I}_i)
-
\mathcal{S}(I_i)
\right|,
\label{eq:boundary_loss}
\end{equation}

where $\mathcal{T}$ denotes the set of target views, $I_i$ and $\hat{I}_i$ denote the ground-truth and rendered target images, respectively. As thermal frames are stored as replicated three-channel grayscale, this computation reduces to a scale-invariant comparison of the single underlying thermal intensity, with no dependence on colour.

\paragraph{Census loss.}

Thermal cameras often produce view-dependent intensity variations due to sensor characteristics and automatic image normalisation. These changes are largely unrelated to scene geometry yet are penalised by pixel-wise photometric losses. To provide supervision that is robust to such intensity variations, we employ a differentiable soft census loss based on the descriptor of~\cite{meister2018unflow,zabih1994non}, which represents each pixel using the relative intensity ordering within its local neighbourhood instead of absolute intensity values.

For each pixel $p$ and neighbour $q\in\mathcal{N}(p)$ in a $k\times k$ window, the descriptor is

\begin{equation}
\xi(I)_{p,q}
=
\frac{L_q-L_p}
{\sqrt{0.81+(L_q-L_p)^2}},
\label{eq:census_desc}
\end{equation}

where $L$ is the grayscale luminance and the constant $0.81$ following~\cite{meister2018unflow}, controls the soft-thresholding scale. This is a differentiable relaxation of the binary census transform~\cite{zabih1994non}, mapping each pairwise comparison to $[-1,1]$ according to its sign and magnitude.

The loss is the average $\ell_1$ distance between the descriptor fields of rendered and ground-truth views:

\begin{equation}
\mathcal{L}_{\text{census}}
=
\frac{1}{|\mathcal{T}|}
\sum_{i\in\mathcal{T}}
\frac{1}{|\Omega|\,k^2}
\sum_{p\in\Omega}
\sum_{q\in\mathcal{N}(p)}
\left|
\xi(\hat{I}_i)_{p,q}
-
\xi(I_i)_{p,q}
\right|,
\label{eq:census}
\end{equation}

where $\Omega$ is the image domain. Because the descriptor is computed from local intensity differences, it is inherently insensitive to uniform brightness shifts and substantially more robust to global intensity remapping than pixel-wise photometric losses. Together, the boundary structure loss and the census loss provide complementary structural supervision: the former encourages sharp and geometrically consistent thermal boundaries, while the latter preserves local intensity relationships despite view-dependent intensity variations. Their combination yields the best reconstruction quality in our experiments (Sec.~\ref{sec:experiments}).



\section{Experiments}
\label{sec:experiments}

\subsection{Dataset}

Since no single thermal dataset provides sufficient scale and diversity to train a generalizable feed-forward reconstruction model, we construct an aggregated thermal corpus by combining five publicly available multi-view thermal datasets: MVTV~\cite{xu2024leveraging}, ThermalGaussian~\cite{lu2024thermalgaussian}, Thermal3DGS~\cite{chen2026thermal3d}, ThermalMix~\cite{ozer2024exploring}, and ThermoNeRF~\cite{hassan2025thermonerf}. The resulting corpus spans a wide range of scene types, including indoor objects, outdoor environments, building facades, industrial equipment, low-light scenes, and UAV aerial captures, providing substantial diversity in geometry, viewpoint, and thermal appearance. 

\subsection{Evaluation Protocol}
 
As ThermalSplat addresses a setting for which no prior method exists (generalizable, pose-free, sparse-view novel view synthesis in the thermal infrared modality), our evaluation has two goals: to establish a credible baseline for this new problem, and to validate each component of our design. We evaluate on held-out thermal scenes from the aggregated dataset. \rev{We use a scene-level split of 57 training and 6 test scenes; (see supplementary Table~1), so metrics measure generalisation to unseen scenes. Tables~2 and~3 use different evaluation protocols (see supplementary).} Synthesis quality is reported with PSNR, SSIM~\cite{wang2004image}, and LPIPS~\cite{zhang2018unreasonable}, computed between synthesised and ground-truth thermal views and averaged over all valid sparse context pairs in the test set. Following prior work~\cite{ye2025no}, we report the Area Under the Cumulative Error Curve (AUC) for pose estimation at $5^\circ$, $10^\circ$, and $20^\circ$ thresholds, where the pose error is defined as the maximum of the rotation error and the translation angular error. All experiments use two context views at $224{\times}224$ resolution.
 
\subsection{Baselines}

To comprehensively evaluate our framework, we benchmark against three classes of baselines. First, \emph{scene-specific thermal optimization methods}: state-of-the-art thermal Gaussian splatting models (Thermal3DGS, ThermalGaussian OMMG, and PhysIR) evaluated under both sparse and dense capture settings to demonstrate the limitations of per-scene optimization. Second, \emph{naive cross-modal transfer}: generalizable Gaussian splatting models pretrained on RGB data and applied to thermal inputs zero-shot, measuring how far RGB priors carry without thermal adaptation. Third, \emph{thermally adapted feed-forward baselines}: these same RGB models fine-tuned on our aggregated thermal corpus, which isolates the contribution of domain adaptation independently of our proposed structural losses. Finally, we conduct an extensive ablation of our own design to isolate the contribution of each structural loss.

\subsection{Implementation Details}
\rev{Our method uses NAS3R with a VGGT backbone, pretrained on RE10K and fine-tuned on thermal data to leverage its learned multi-view geometry priors.} We \rev{use} AdamW ($\text{lr}=10^{-5}$, weight decay $0.05$) for 10{,}000 steps at $224{\times}224$, batch size 4, $\lambda_b=\lambda_c=0.075$, \rev{and $\mathcal{L}_{\text{photo}}=\mathcal{L}_{\text{MSE}}+0.5\,\mathcal{L}_{\text{LPIPS}}$,} on an RTX Ada 6000 GPU.


\begin{table*}[t]
\centering
\caption{Quantitative comparison against scene-specific optimization methods. We evaluate baselines under their ideal Dense multi-view setting (top) and under the heavily constrained Sparse two-view setting (bottom). ThermalSplat (Ours) operates strictly from two unposed views, outperforming baselines in the sparse regime and rivaling them in the dense regime. Best results per setting are shown in \textbf{bold}.}
\label{tab:thermal_comparison_combined}
\resizebox{\textwidth}{!}{
\begin{tabular}{l|ccc|ccc|ccc|ccc|ccc|ccc}
\toprule
\multirow{2}{*}{Method} 
& \multicolumn{3}{c|}{Hand} 
& \multicolumn{3}{c|}{Dark Scenes} 
& \multicolumn{3}{c|}{Fire3Raw} 
& \multicolumn{3}{c|}{BuildingA Winter} 
& \multicolumn{3}{c|}{Transmission Tower} 
& \multicolumn{3}{c}{Mean}\\
\cmidrule(lr){2-19}
& PSNR$\uparrow$ & SSIM$\uparrow$ & LPIPS$\downarrow$ 
& PSNR$\uparrow$ & SSIM$\uparrow$ & LPIPS$\downarrow$ 
& PSNR$\uparrow$ & SSIM$\uparrow$ & LPIPS$\downarrow$ 
& PSNR$\uparrow$ & SSIM$\uparrow$ & LPIPS$\downarrow$ 
& PSNR$\uparrow$ & SSIM$\uparrow$ & LPIPS$\downarrow$ 
& PSNR$\uparrow$ & SSIM$\uparrow$ & LPIPS$\downarrow$\\
\midrule
\multicolumn{19}{c}{\textit{Dense Multi-View Input (Scene-Specific Optimization Baseline)}} \\
\midrule
Thermal3DGS
& \textbf{33.010} & \textbf{0.9375} & \textbf{0.1558}
& 17.370 & 0.7321 & 0.4568
& 29.870 & 0.9734 & 0.1503
& 32.420 & 0.9576 & 0.1759
& 17.790 & 0.6988 & 0.3311
& 25.140 & 0.8383 & 0.2522\\
ThermalGaussian$^*$
& 30.110 & 0.9210 & 0.1724
& 18.260 & \textbf{0.7798} & 0.4250
& 30.410 & 0.9789 & 0.1373
& \textbf{33.240} & \textbf{0.9659} & \textbf{0.1545}
& 18.060 & 0.7013 & 0.3162
& 26.020 & \textbf{0.8694} & \textbf{0.2411}\\
PhysIR
& 31.050 & 0.9290 & 0.1640
& \textbf{18.490} & 0.7600 & \textbf{0.4220}
& \textbf{35.190} & \textbf{0.9800} & \textbf{0.1357}
& 30.710 & 0.9400 & 0.2100
& \textbf{18.390} & \textbf{0.7095} & \textbf{0.3100}
& \textbf{26.766} & 0.8622 & 0.2773\\
\midrule
\multicolumn{19}{c}{\textit{Sparse Two-View Input (Constraint Regime)}} \\
\midrule
Thermal3DGS
& 20.591 & 0.8432 & 0.2705
& 15.208 & 0.6869 & 0.5602
& 5.873 & 0.0010 & 0.5249
& 10.466 & 0.5350 & 0.7086
& 7.305 & 0.3775 & 0.6705
& 11.889 & 0.4887 & 0.5469\\
ThermalGaussian$^*$
& 17.628 & 0.7237 & 0.4192
& 19.305 & \textbf{0.7732} & 0.4348
& 15.547 & 0.6580 & 0.2445
& 17.745 & 0.7550 & 0.5158
& 5.581 & 0.3004 & 0.6734
& 15.161 & 0.6420 & 0.4583\\
PhysIR
& 24.360 & 0.8594 & 0.2119
& 13.588 & 0.6101 & 0.6215
& \textbf{33.870} & \textbf{0.9788} & \textbf{0.1376}
& 5.769 & 0.4921 & 0.8322
& \textbf{17.539} & \textbf{0.6339} & 0.2920
& 19.025 & 0.7149 & 0.4190\\
\addlinespace
\textbf{Ours (Sparse)}
& \textbf{28.410} & \textbf{0.8900} & \textbf{0.1020}
& \textbf{19.440} & 0.7320 & \textbf{0.3800}
& 30.040 & 0.9710 & 0.1570
& \textbf{29.830} & \textbf{0.9420} & \textbf{0.1290}
& 17.100 & 0.5620 & \textbf{0.2780}
& \textbf{24.964} & \textbf{0.8194} & \textbf{0.2092}\\
\bottomrule
\end{tabular}}
{\raggedright\scriptsize $^*$ The UAV6 scene was excluded as it lacks corresponding RGB data required by ThermalGaussian. For fair comparison, it was excluded across all methods.\par}
\end{table*}

\begin{table*}[!t]
\centering
\caption{Quantitative comparison of novel view synthesis performance across six held-out thermal scenes. In the absence of existing generalizable frameworks for sparse-view thermal reconstruction, we benchmark against state-of-the-art RGB models. \rev{We first evaluate these models using zero-shot inference and, for a fair comparison, also fine-tune them on our aggregated thermal dataset}. \rev{The} fine-tuned versions are denoted with a superscript plus ($^+$).}
\label{tab:baseline_comparison}
\resizebox{\textwidth}{!}{%
\begin{tabular}{l|ccc|ccc|ccc|ccc|ccc|ccc|ccc}
\hline
\multirow{2}{*}{Method}
& \multicolumn{3}{c|}{Hand}
& \multicolumn{3}{c|}{Dark Scenes}
& \multicolumn{3}{c|}{Fire3Raw}
& \multicolumn{3}{c|}{BuildingA Winter}
& \multicolumn{3}{c|}{Transmission Tower}
& \multicolumn{3}{c|}{UAV6}
& \multicolumn{3}{c}{Mean} \\
\cline{2-22}
& PSNR$\uparrow$ & SSIM$\uparrow$ & LPIPS$\downarrow$
& PSNR$\uparrow$ & SSIM$\uparrow$ & LPIPS$\downarrow$
& PSNR$\uparrow$ & SSIM$\uparrow$ & LPIPS$\downarrow$
& PSNR$\uparrow$ & SSIM$\uparrow$ & LPIPS$\downarrow$
& PSNR$\uparrow$ & SSIM$\uparrow$ & LPIPS$\downarrow$
& PSNR$\uparrow$ & SSIM$\uparrow$ & LPIPS$\downarrow$
& PSNR$\uparrow$ & SSIM$\uparrow$ & LPIPS$\downarrow$ \\
\hline
MVSplat
& 10.88 & 0.150 & 0.785
& 3.65  & 0.006 & 0.860
& 6.60  & 0.009 & 0.731
& 6.36  & 0.137 & 0.882
& 6.22  & 0.153 & 0.713
& 15.21 & 0.432 & 0.510
& 8.15  & 0.148 & 0.747 \\
MVSplat$^+$
& 12.41 & 0.475 & 0.594
& 9.68  & 0.454 & 0.762
& 23.91 & 0.880 & 0.296
& 23.20 & 0.840 & 0.341
& 14.13 & 0.450 & 0.375
& 18.49 & 0.513 & 0.367
& 16.97 & 0.602 & 0.456 \\
SPFSplat
& 18.03 & 0.681 & 0.313
& 18.46 & 0.718 & 0.434
& \textbf{32.26} & 0.953 & 0.216
& 13.16 & 0.677 & 0.525
& 16.72 & 0.599 & 0.299
& 12.13 & 0.278 & 0.645
& 18.46 & 0.651 & 0.405 \\
SPFSplat$^+$
& 18.38 & 0.691 & 0.374
& 19.40 & 0.718 & 0.467
& 29.42 & 0.962 & 0.201
& 20.66 & 0.815 & 0.464
& 13.30 & 0.677 & 0.534
& 14.64 & 0.337 & 0.517
& 19.30 & 0.700 & 0.426 \\
NAS3R
& \textbf{29.33} & \textbf{0.918} & \textbf{0.085}
& 18.20 & 0.702 & 0.402
& 31.06 & 0.958 & 0.200
& 24.78 & 0.882 & 0.221
& \textbf{18.01} & \textbf{0.626} & \textbf{0.249}
& 19.26 & 0.523 & 0.281
& 23.44 & 0.768 & 0.240 \\
NAS3R$^+$
& 25.39 & 0.827 & 0.119
& 19.81 & 0.747 & 0.359
& 32.24 & 0.965 & 0.169
& 27.53 & 0.876 & 0.229
& 16.60 & 0.519 & 0.307
& 24.46 & 0.776 & 0.135
& 24.34 & 0.785 & 0.220 \\
\hline
ThermalSplat(Ours)
& 28.01 & 0.890 & 0.101
& \textbf{20.23} & \textbf{0.745} & \textbf{0.347}
& 29.31 & \textbf{0.969} & \textbf{0.163}
& \textbf{30.21} & \textbf{0.943} & \textbf{0.124}
& 17.49 & 0.573 & 0.278
& \textbf{28.02} & \textbf{0.849} & \textbf{0.070}
& \textbf{25.55} & \textbf{0.828} & \textbf{0.181} \\
\hline
\end{tabular}%
}
\end{table*}

\begin{table}[t]
\centering
\caption{Two-view pose estimation performance measured by the Area Under the Curve (AUC, \%) at error thresholds of $5^\circ$, $10^\circ$, and $20^\circ$. We report rotation, translation, and overall errors. For each image pair, the overall error is simply the larger of its rotation or translation error.}
\label{tab:pose}
\setlength{\tabcolsep}{3pt}
\renewcommand{\arraystretch}{1.1}
\resizebox{\columnwidth}{!}{%
\begin{tabular}{@{}lrrrrrrrrr@{}}
\toprule
\multirow{2}{*}{\textbf{Method}}
  & \multicolumn{3}{c}{Rotation}
  & \multicolumn{3}{c}{Translation}
  & \multicolumn{3}{c}{Overall} \\
\cmidrule(l){2-4}\cmidrule(lr){5-7}\cmidrule(l){8-10}
  & $5^\circ{\uparrow}$ & $10^\circ{\uparrow}$ & $20^\circ{\uparrow}$
  & $5^\circ{\uparrow}$ & $10^\circ{\uparrow}$ & $20^\circ{\uparrow}$
  & $5^\circ{\uparrow}$ & $10^\circ{\uparrow}$ & $20^\circ{\uparrow}$ \\
\midrule
NAS3R
  & 43.2 & 55.2 & 70.0
  & 10.5 & 23.2 & 37.7
  &  9.5 & 21.3 & 35.8 \\
NAS3R$^+$
  & 51.3 & \textbf{68.0} & \textbf{80.7}
  &  9.5 & 23.4 & 40.2
  &  7.7 & 21.5 & 38.6 \\
\textbf{ThermalSplat (Ours)}
  & \textbf{51.5} & 67.9 & 80.5
  & \textbf{11.9} & \textbf{25.7} & \textbf{41.2}
  & \textbf{10.4} & \textbf{23.8} & \textbf{39.7} \\
\bottomrule
\end{tabular}%
}
\end{table}

\begin{table}[t]
\centering
\small
\caption{Per-scene reconstruction time comparison. Optimization baselines require lengthy training on dense views, whereas ThermalSplat infers the scene in a single forward pass.}
\label{tab:runtime}
\begin{tabular}{@{}llr@{}}
\toprule
\textbf{Method} & \textbf{Paradigm} & \textbf{Time} $\downarrow$ \\
\midrule
Thermal3DGS          & Per-scene Opt. & $\sim$15 min \\
ThermalGaussian OMMG & Per-scene Opt. & $\sim$25 min \\
PhysIR               & Per-scene Opt. & $\sim$15 min \\
\midrule
\textbf{ThermalSplat (Ours)} & \textbf{Feed-Forward} & \textbf{80 ms} \\
\bottomrule
\end{tabular}
\end{table}

\begin{table}[t]
\centering
\caption{Ablation study evaluating the contribution of each proposed loss component. Reference rows establish baselines using an RGB-pretrained backbone applied zero-shot (Naive transfer) and fine-tuned with thermal photometric supervision .}
\label{tab:ablation}
\small
\setlength{\tabcolsep}{4pt}
\begin{tabular}{lccc}
\toprule
\textbf{Variant} & \textbf{PSNR $\uparrow$} &
\textbf{SSIM $\uparrow$} & \textbf{LPIPS $\downarrow$} \\
\midrule
\multicolumn{4}{l}{\textit{Reference: RGB-pretrained \& Fine-tuned baselines}} \\
\midrule
Naive transfer & 23.44 & 0.768 & 0.240 \\
Photometric    & 24.34 & 0.785 & 0.220 \\
\midrule
\multicolumn{4}{l}{\textit{ThermalSplat ablations}} \\
\midrule
+ Census Loss       & 25.15 & 0.810 & 0.192 \\
+ Boundary Loss  & 25.46 & 0.822 & 0.183 \\
\textbf{Boundary + Census} & \textbf{25.55} & \textbf{0.828} & \textbf{0.181} \\
\bottomrule
\end{tabular}
\end{table}


\subsection{Quantitative Results}

\begin{figure*}[t]
    \centering
    \includegraphics[width=0.9\textwidth]{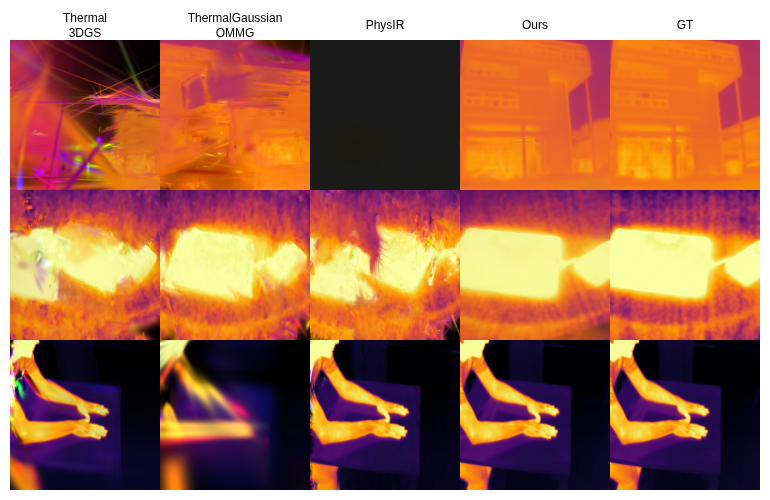}
    \vspace{-3mm}
    \caption{Qualitative comparison under the sparse two-view setting against scene-specific optimization-based thermal reconstruction methods. When restricted to highly sparse inputs, existing thermal baselines suffer catastrophic geometry collapse, whereas ThermalSplat (Ours) maintains sharp boundary delineation and structural integrity. For comparisons under the dense-view setting, please refer to the supplementary material.}
    \label{fig:qual_thermal_comp_sparse}
\end{figure*}

\paragraph{Optimization methods degrade severely under extreme sparsity.}
When restricted to the identical sparse two-view input setting (Table~\ref{tab:thermal_comparison_combined}, bottom), scene-specific baselines experience severe degradation. The mean PSNR for Thermal3DGS and PhysIR drops precipitously to \num{11.889} and \num{19.025}. In sharp contrast, ThermalSplat leverages learned geometric priors to maintain high-fidelity reconstructions (\rev{\num{24.964}} PSNR, \rev{\num{0.8194}} SSIM, \rev{\num{0.2092}} LPIPS), confirming the fundamental necessity of a feed-forward architecture for sparse thermal inputs.
\vspace{-4mm}
\paragraph{Sparse pose-free inference rivals dense per-scene optimization.}
We further benchmark ThermalSplat against these baselines operating under their ideal dense capture settings with ground-truth poses (Table~\ref{tab:thermal_comparison_combined}, top). Despite generalizing from only two unposed views, ThermalSplat closely rivals optimization-based baselines in PSNR (\num{24.96}) and decisively outperforms all of them in perceptual quality. Our mean LPIPS of \num{0.2092} represents substantial improvements of \num{17.1}\%, \num{13.2}\%, and \num{24.5}\% over Thermal3DGS, ThermalGaussian OMMG, and PhysIR, respectively, effectively bypassing the need for dense multi-view capture. 
\vspace{-4mm}
\paragraph{Cross-Dataset Generalization.}
ThermalSplat generalizes well to the unseen SEAR~\cite{skorokhodov2026sear} dataset, achieving strong results across six scenes and outperforming PhysIR-Splat on average (see supplementary Table~4).
\vspace{-4mm}
\paragraph{Feed-forward inference eliminates per-scene optimization.}
Beyond reconstruction quality, ThermalSplat offers massive efficiency gains. While optimization-based methods require 15 to 25 minutes of per-scene training (Table~\ref{tab:runtime}), ThermalSplat reconstructs scenes in a single \num{80}-millisecond forward pass, enabling time-sensitive, real-world deployments.
\vspace{-4mm}
\paragraph{Thermal adaptation and structural losses drive perceptual fidelity.}
Comparing our framework against generalizable RGB models (Table~\ref{tab:baseline_comparison}), the strongest zero-shot baseline (NAS3R) confirms that multi-view RGB priors partially survive the modality gap (\num{23.44} PSNR) but fall severely short in perceptual quality (\num{0.240} LPIPS). While thermal fine-tuning (NAS3R$^+$) improves structural fidelity, ThermalSplat pushes performance significantly higher by integrating our boundary and census losses. We reach a mean PSNR of \num{25.55} and drop LPIPS to \num{0.181}, achieving the high perceptual fidelity that is critical for operational thermal analysis.
 \vspace{-4mm}
\paragraph{Targeted structural losses foster robust multi-view geometry.}
Superior appearance supervision translates directly into more accurate underlying geometry from unposed inputs. As detailed in Table~\ref{tab:pose}, ThermalSplat systematically outperforms zero-shot and fine-tuned baselines, improving the overall pose estimation AUC to \num{39.7}\% at the $20^\circ$ threshold by anchoring representations on consistent temperature gradients.

\subsection{Qualitative Analysis}

\begin{figure*}[t]
    \centering
    \includegraphics[width=0.9\textwidth]{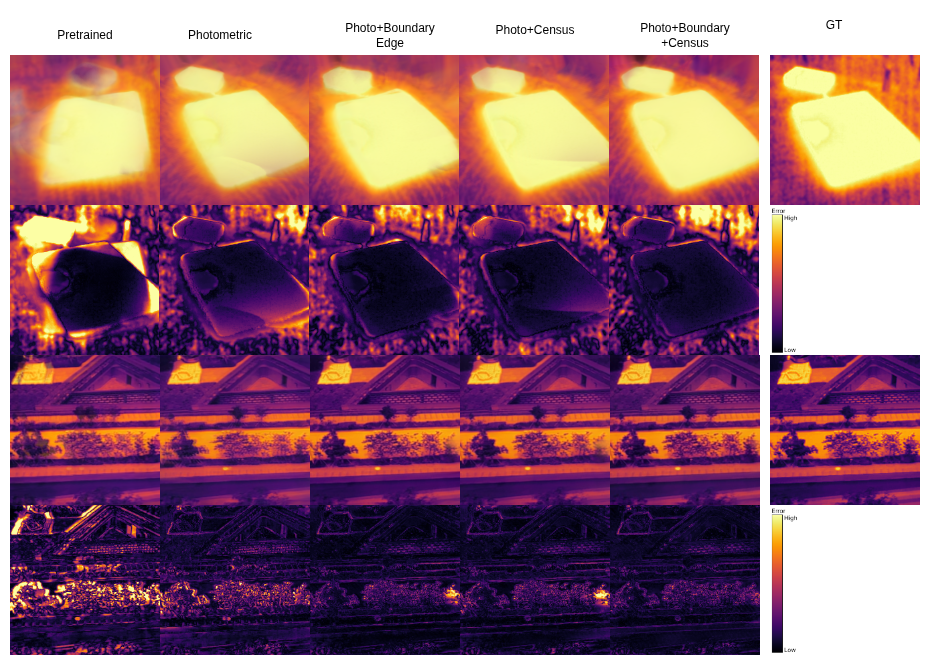}
    \vspace{-3mm}
    \caption{Qualitative ablation study results showing novel view synthesis and the corresponding error maps ($\Delta$GT). The full ThermalSplat model (second-from-right column) minimizes structural artifacts and boundary blurring compared to partial configurations, confirming the synergy between the boundary structure and soft census losses.}
    \label{fig:qual_ablation}
\end{figure*}

Visual comparisons corroborate our quantitative findings. Under sparse two-view constraints (Figure~\ref{fig:qual_thermal_comp_sparse}), scene-specific optimization models like Thermal3DGS and PhysIR suffer catastrophic geometry collapse, producing severe uninterpretable ghosting or entirely black outputs (e.g., in the Dark Scenes sequence). Conversely, ThermalSplat maintains sharp boundary delineation and structural integrity, far exceeding these degraded baselines.

Furthermore, as detailed in the supplementary material, our sparse-view framework rivals the visual fidelity of these baselines even when they operate on ideal dense captures. Our boundary structure and census losses effectively suppress view-dependent intensity shifts to yield crisper novel views. The supplement also illustrates the inadequacy of simple cross-modal transfer; zero-shot and thermally fine-tuned RGB baselines suffer from intensity collapse and blurred edges, whereas ThermalSplat successfully anchors scene geometry using sharp temperature gradients.

\subsection{Ablation Study}

\paragraph{Complementary structural losses drive perceptual fidelity.}
To isolate the contribution of our specific architectural choices, we conduct an extensive ablation study. As detailed in Table~\ref{tab:ablation}, relative to the standard photometric-only model, both proposed structural losses consistently improve synthesis quality. Their combination yields the best overall result: ThermalSplat (full) reaches \textbf{\num{25.55}} PSNR, \textbf{\num{0.828}} SSIM, and \textbf{\num{0.181}} LPIPS. The combined model attains the strongest SSIM and LPIPS of any configuration, confirming that the two losses are complementary. The boundary structure loss supervises the location of temperature boundaries, while the census loss enforces brightness-invariant local structure. Furthermore, qualitative error maps ($\Delta$GT) provided in Figure~\ref{fig:qual_ablation} clearly illustrate that perceptual accuracy is maximized and structural artifacts are minimized only when both complementary loss components are active during training.
\section{Discussion and Conclusion}
\label{sec:discussion_and_conclusion}

\paragraph{Limitations and Future Work}
While ThermalSplat establishes a new state-of-the-art for pose-free, sparse-view thermal reconstruction, it faces specific limitations that open promising avenues for future research. \textbf{Dynamic environments:} Our method assumes static scenes. Future work will incorporate temporal attention or deformation fields into the transformer backbone to handle moving heat signatures in applications like search-and-rescue. \textbf{Geometric ambiguity in unobserved regions for wide baseline captures:} Relying on \rev{few} sparse views leaves large volumes unobserved for wide baseline captures.
\vspace{-4mm} 
\paragraph{Conclusion}
We have presented ThermalSplat, a pose-free generalizable, feed-forward Gaussian splatting framework for sparse-view thermal 3D reconstruction. By fine-tuning a large pose-free geometry backbone on an aggregated multi-source thermal corpus and supervising the reconstruction with thermal adapted structural loss terms, specifically a boundary loss and a soft census loss tailored to the emissivity and low-texture characteristics of thermal imagery, ThermalSplat consistently outperforms both zero-shot and thermally adapted RGB baselines. Furthermore, despite operating from only two unposed views in a single forward pass without RGB guidance, our method achieves a visual fidelity that rivals and often exceeds the perceptual quality of dense, scene-specific optimization frameworks. To the best of our knowledge, ThermalSplat is the first generalizable, pose-free framework for this problem, establishing a strong standard at the intersection of thermal imaging and scene representation.

\paragraph{Acknowledgements}
KM would like to acknowledge support from Qualcomm Faculty Award 2024.

{
    \small
    \bibliographystyle{ieeenat_fullname}
    \bibliography{main}
}

\end{document}


\maketitle

\section{Dataset Composition and Split}
To train a robust, generalizable thermal reconstruction model, we aggregate five publicly available multi-view thermal datasets.\rev{All five datasets contain non-radiometric thermal images with 8-bit depth.}This corpus spans diverse indoor and outdoor environments, ranging from object-centric captures to large-scale scenes under varying imaging conditions. The aggregated dataset comprises 63 distinct scenes, partitioned strictly at the scene level into 57 training and 6 test scenes. This scene-level split ensures no geometric or appearance leakage occurs between subsets, guaranteeing a rigorous evaluation of the model's ability to generalize to entirely unseen environments. Table~\ref{tab:dataset_split} details the source datasets and their respective splits.

\newcommand{\testscene}{\textsuperscript{\textbf{T}}}

\begin{table*}[!htbp]
\centering
\small
\caption{Composition of the aggregated thermal dataset constructed from five publicly available datasets. The number in parentheses next to each dataset name indicates the total number of scenes contributed by that dataset. The complete corpus contains 63 scenes, partitioned into 57 training (unmarked) and 6 test (\textbf{bold}) scenes.}
\label{tab:dataset_split}
\begin{tabular}{lp{12.5cm}}
\toprule
\textbf{Source} & \textbf{Scenes used} \\
\midrule
ThermalGaussian (14) &
building, RotaryKiln, ebike, parterre, Plant\_Equipment, daily\_stuff,
Glass\_Cup, Dimsum, RoadBlock, ironvogt, landscape, Truck,
\textbf{Dark\_Scenes}, \textbf{Transmission\_tower} \\
\addlinespace

Thermal3DGS (17) &
heated, corridor, bridge, wall, chair, sitting, standing\_tall,
building, bicycle, soccer\_goal, merge, UAV3, UAV4, apple,
UAV5, \textbf{UAV6}, UAV1 \\
\addlinespace

ThermoNeRF (9) &
exhibition\_building, trees, melting\_ice\_cup, double\_robot,
heater\_watercup, heater\_water\_kettle,
BuildingA\_spring, \textbf{BuildingA\_winter}, raspberry \\
\addlinespace

MVTV (18) &
human0\_raw, human1\_raw, human2\_raw, fire0\_raw, fire1\_raw,
fire2\_raw, car0\_raw, car1\_raw, bench\_raw, tree\_raw,
corner\_raw, road\_raw, chair\_raw, grillman\_raw, mason\_raw,
forest\_raw, \textbf{fire3\_raw}, human\_raw \\
\addlinespace

ThermalMix (5) &
face, laptop, pan, pan1, \textbf{hand} \\
\midrule
\textbf{Total} & \textbf{63 scenes} \\
\bottomrule
\end{tabular}
\end{table*}

\section{Comparison with Optimization-Based Thermal Methods}
\begin{figure*}[!b]
    \centering
    \includegraphics[width=\textwidth]{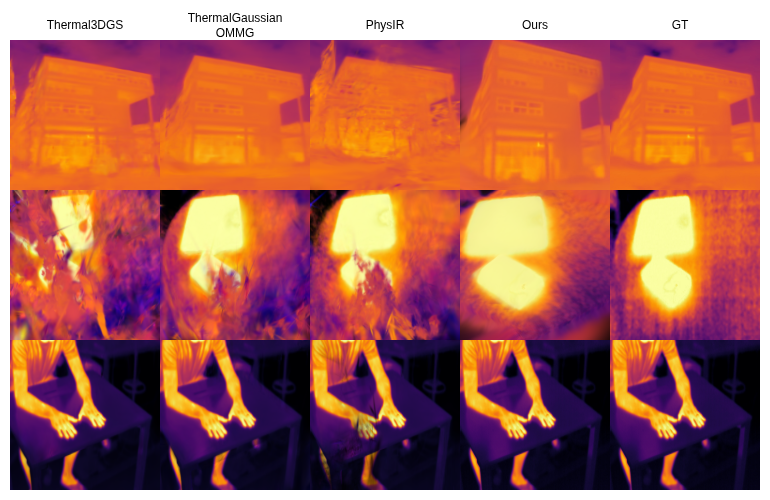}
    \caption{Qualitative comparison against scene-specific, optimization-based thermal reconstruction methods with dense input views. Despite generalizing from only two unposed views in a single feed-forward pass without RGB guidance, ThermalSplat (Ours) achieves visual fidelity and boundary sharpness that rivals or exceeds heavily optimized, dense multi-view baselines (Thermal3DGS, ThermalGaussian, and PhysIR).}
    \label{fig:qual_thermal_comp_dense}
\end{figure*}

Figure~\ref{fig:qual_thermal_comp_dense} contrasts ThermalSplat with state-of-the-art thermal scene-specific optimization methods, including Thermal3DGS, ThermalGaussian, and PhysIR. While these baselines require calibrated camera poses, dense multi-view observations, and extensive per-scene optimization, ThermalSplat achieves comparable novel view synthesis from only two unposed input images in a single feed-forward pass. To ensure a strictly fair quantitative comparison against these per-scene optimization models (as presented in Table 2 of the main manuscript), we evaluate and average our metrics exclusively over their designated held-out test views.  This contrasts with our generalizable benchmarking (Table 3 of the main manuscript), which \rev{evaluates} synthesis quality across all available views within \rev{held-out} test scenes. Ultimately, the visual evidence demonstrates that our method faithfully reconstructs scene geometry and preserves sharp thermal boundaries, matching the visual fidelity of computationally heavy optimization pipelines at a fraction of the informational and computational cost.

\section{Qualitative Comparison with Generalizable Baselines}
\begin{figure*}[!t]
    \centering
    \includegraphics[width=\textwidth]{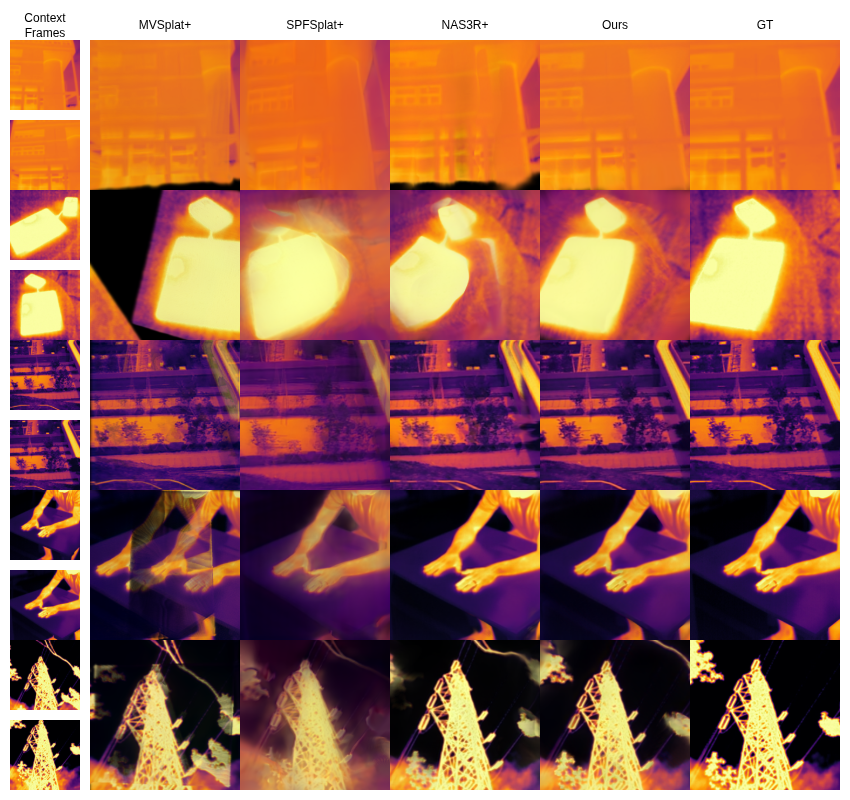}
    \caption{Qualitative comparison of novel view synthesis. ThermalSplat (Ours) captures sharper temperature discontinuities and more faithful structural details compared to thermally fine-tuned baselines (MVSplat$^+$, SPFSplat$^+$, NAS3R$^+$).}
    \label{fig:qual_baselines}
\end{figure*}

Figure~\ref{fig:qual_baselines} provides additional qualitative comparisons against recent generalizable sparse-view reconstruction architectures. We benchmark against thermally fine-tuned variants of MVSplat, SPFSplat, and NAS3R. Across a diverse array of indoor and outdoor scenes, ThermalSplat consistently reconstructs sharper object boundaries and produces more accurate thermal appearance. Notably, ThermalSplat preserves fine thermal boundaries and fine structures than competing methods.

\section{Scaling to More Input Views}
\label{sec:view_scaling}
ThermalSplat readily scales to multiple context views without architectural changes. Training and evaluating the model across different multi-view configurations (Table~\ref{tab:view_scaling}) demonstrates that supplying additional views consistently improves performance. Increasing the context from two to five views elevates average PSNR from 25.55\,dB to 26.35\,dB, while simultaneously improving SSIM and LPIPS. This confirms our framework effectively leverages richer context to refine both geometric structure and thermal appearance.

\begin{table}[H]
    \centering
    \caption{Reconstruction performance across different multi-view context configurations.}
    \label{tab:view_scaling}
    \begin{tabular}{cccc}
    \toprule
    \textbf{\# Views} & \textbf{PSNR} $\uparrow$ & \textbf{SSIM} $\uparrow$ & \textbf{LPIPS} $\downarrow$ \\
    \midrule
    2 & 25.55 & 0.828 & 0.181 \\
    3 & 25.83 & 0.827 & 0.186 \\
    5 & \textbf{26.35} & \textbf{0.829} & \textbf{0.179} \\
    \bottomrule
    \end{tabular}
\end{table}


\section{Sensitivity Analysis of Loss Weights}
\label{sec:sensitivity}
We evaluate the sensitivity of ThermalSplat to the boundary ($\lambda_b$) and census ($\lambda_c$) loss weights. As shown in Table~\ref{tab:lambda_sensitivity}, the balanced setting $\lambda_b=\lambda_c=0.075$ achieves the best overall performance and is therefore used in all experiments.

\begin{table*}[!t] 
    \centering
    \caption{Sensitivity analysis of the boundary loss weight ($\lambda_b$) and census loss weight ($\lambda_c$). Results are reported for each evaluation scene and the overall mean. Best values are highlighted in \textbf{bold}.}
    \label{tab:lambda_sensitivity}
    \resizebox{\textwidth}{!}{%
    \begin{tabular}{cc|ccc|ccc|ccc|ccc|ccc|ccc|ccc}
    \toprule
    \multirow{2}{*}{$\lambda_b$} &
    \multirow{2}{*}{$\lambda_c$} &
    \multicolumn{3}{c|}{Hand} &
    \multicolumn{3}{c|}{Dark Scenes} &
    \multicolumn{3}{c|}{Fire3Raw} &
    \multicolumn{3}{c|}{BuildingA\_winter} &
    \multicolumn{3}{c|}{Transmission Tower} &
    \multicolumn{3}{c|}{UAV6} &
    \multicolumn{3}{c}{Mean} \\
    \cmidrule(lr){3-5} \cmidrule(lr){6-8} \cmidrule(lr){9-11} \cmidrule(lr){12-14} \cmidrule(lr){15-17} \cmidrule(lr){18-20} \cmidrule(lr){21-23}
    & & PSNR & SSIM & LPIPS & PSNR & SSIM & LPIPS & PSNR & SSIM & LPIPS & PSNR & SSIM & LPIPS & PSNR & SSIM & LPIPS & PSNR & SSIM & LPIPS & PSNR & SSIM & LPIPS \\
    \midrule
    0.05 & 0.05 &
    27.90 & 0.889 & 0.103 & 19.83 & 0.741 & 0.354 & \textbf{29.50} & 0.968 & \textbf{0.160} & 29.91 & 0.941 & 0.127 & 17.47 & 0.566 & \textbf{0.273} & 27.01 & 0.850 & \textbf{0.069} & 25.44 & 0.8258 & 0.1810 \\
    0.05 & 0.10 &
    27.28 & \textbf{0.893} & 0.102 & 20.00 & 0.747 & 0.350 & 29.85 & 0.968 & \textbf{0.160} & 29.94 & 0.940 & 0.130 & 17.05 & 0.547 & 0.292 & 28.24 & \textbf{0.855} & 0.071 & 25.39 & 0.8250 & 0.1842 \\
    0.10 & 0.05 &
    27.73 & \textbf{0.893} & 0.102 & \textbf{20.26} & \textbf{0.750} & \textbf{0.336} & 29.37 & 0.968 & 0.163 & 29.95 & 0.942 & 0.125 & 17.20 & 0.550 & 0.286 & \textbf{28.28} & 0.854 & 0.075 & 25.47 & 0.8262 & 0.1808 \\
    0.075 & 0.075 &
    \textbf{28.01} & 0.890 & \textbf{0.101} & 20.23 & 0.745 & 0.347 & 29.31 & \textbf{0.969} & 0.163 & \textbf{30.21} & \textbf{0.943} & \textbf{0.124} & \textbf{17.49} & \textbf{0.573} & 0.278 & 28.02 & 0.849 & 0.070 & \textbf{25.55} & \textbf{0.8282} & \textbf{0.1805} \\
    \bottomrule
    \end{tabular}
    }
\end{table*}

\begin{table*}[t]
\centering
\caption{Cross-dataset generalization results on the SEAR dataset. Best results are highlighted in bold.}
\label{tab:sear_generalization}
\resizebox{\textwidth}{!}{
\begin{tabular}{l|ccc|ccc}
\toprule
\multirow{2}{*}{Scene}
& \multicolumn{3}{c|}{PhysIR-Splat}
& \multicolumn{3}{c}{ThermalSplat} \\
& PSNR $\uparrow$ & SSIM $\uparrow$ & LPIPS $\downarrow$
& PSNR $\uparrow$ & SSIM $\uparrow$ & LPIPS $\downarrow$ \\
\midrule
Statue
& 29.19 & 0.938 & 0.212
& \textbf{33.22} & \textbf{0.962} & 0.131 \\
Telescope
& 31.68 & \textbf{0.940} & \textbf{0.088}
& \textbf{32.12} & 0.936 & 0.097 \\
Old-drinking-fountain
& \textbf{35.37} & \textbf{0.968} & 0.213
& 35.11 & 0.967 & \textbf{0.129} \\
Parking
& \textbf{36.50} & \textbf{0.981} & 0.107
& 36.45 & 0.972 & \textbf{0.097} \\
Metallic-container
& 7.61 & 0.370 & 0.370
& \textbf{33.18} & \textbf{0.969} & \textbf{0.137} \\
Conference-room
& 8.98 & 0.414 & \textbf{0.164}
& \textbf{27.08} & \textbf{0.943} & 0.187 \\
\midrule
Mean (6 scenes)
& 24.89 & 0.768 & 0.192
& \textbf{32.86} & \textbf{0.958} & \textbf{0.130} \\
\bottomrule
\end{tabular}
}
\end{table*}


\rev{\section{Cross-Dataset Generalization on the SEAR Dataset}}
\label{sec:sear}

\begin{figure*}[!t]
    \centering
    \includegraphics[width=\textwidth]{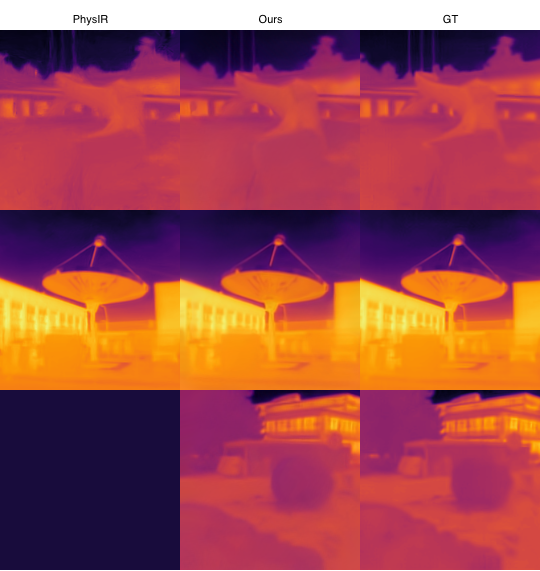}
    \caption{\rev{Qualitative comparison on the unseen SEAR dataset. PhysIR-Splat collapses on the Metallic-container scene, producing a uniform output, whereas ThermalSplat (Ours) preserves sharp thermal boundaries and scene structure across all scenes.}}
    \label{fig:qual_sear}
\end{figure*}

\rev{To evaluate cross-dataset generalization, we test ThermalSplat on the unseen SEAR dataset. As shown in Table~\ref{tab:sear_generalization}, ThermalSplat achieves an average PSNR of 32.86\,dB, SSIM of 0.958, and LPIPS of 0.130, demonstrating its strong generalization to unseen thermal scenes. Figure~\ref{fig:qual_sear} shows representative novel views. On scenes where PhysIR-Splat converges (Statue, Telescope), both methods recover the scene structure, while on Metallic-container PhysIR-Splat collapses to a uniform output (7.61\,dB PSNR), whereas ThermalSplat reconstructs the scene faithfully.}

